\documentclass{amia}
\usepackage{graphicx}
\usepackage[labelfont=bf]{caption}
\usepackage[superscript,nomove]{cite}
\usepackage{color}
\usepackage{caption} 
\usepackage[
singlelinecheck=false 
]{caption}
\usepackage{titlesec}
\titlespacing*{\section}{0pt}{*0}{*0}
\usepackage{enumitem}
\usepackage{multicol}
\usepackage[para,symbol]{footmisc}  
\usepackage{soul}
\usepackage[color=yellow, textsize=small]{todonotes}
\usepackage{booktabs}
\usepackage[hidelinks]{hyperref}
\usepackage{hyperref}
\usepackage{url}
\usepackage{subfiles}
\usepackage{multirow}
\usepackage{subcaption}
\makeatletter
\renewcommand{\@biblabel}[1]{\hfill #1.}
\makeatother
\usepackage[export]{adjustbox}
\usepackage{wrapfig,lipsum,booktabs}
\let\oldbibliography\thebibliography
\renewcommand{\thebibliography}[1]{%
  \oldbibliography{#1}%
  \setlength{\itemsep}{-0.1em}%
}

\usepackage{caption}
\usepackage{multicol}
\usepackage{calc} 
\usepackage[normalem]{ulem}
\useunder{\uline}{\ul}{}

\title{BioMistral-NLU: Towards More Generalizable Medical \\ Language Understanding through Instruction Tuning}

\author{ 
\begin{center}
    Yujuan Velvin Fu, BSE$^1$, Giridhar Kaushik Ramachandran, PhD$^2$, Namu Park, MS$^1$,  \\ Kevin Lybarger, PhD$^2$,  Fei Xia, PhD$^1$,  Ozlem Uzuner, PhD$^2$,  Meliha Yetisgen, PhD$^1$
\end{center}
}
\institutes{ 
\begin{center}
    $^1$University of Washington, Seattle, WA, USA; $^2$ George Mason University, Fairfax, VA, USA
\end{center}
}

\begin{document}
\maketitle

\section*{Abstract}
Large language models (LLMs) such as ChatGPT are fine-tuned on large and diverse instruction-following corpora, and can generalize to new tasks. However, those instruction-tuned LLMs often perform poorly in specialized medical natural language understanding (NLU) tasks that require domain knowledge, granular text comprehension, and structured data extraction.
To bridge the gap, we: (1) propose a unified prompting format for 7 important NLU tasks, (2) curate an instruction-tuning dataset, MNLU-Instruct, utilizing diverse existing open-source medical NLU corpora, and (3) develop BioMistral-NLU, a generalizable medical NLU model, through fine-tuning BioMistral on MNLU-Instruct. We evaluate BioMistral-NLU in a zero-shot setting, across 6 important NLU tasks, from two widely adopted medical NLU benchmarks: BLUE and BLURB. Our experiments show that our BioMistral-NLU outperforms the original BioMistral, as well as the proprietary LLMs - ChatGPT and GPT-4. 
Our dataset-agnostic prompting strategy and instruction tuning step over diverse NLU tasks enhance LLMs' generalizability across diverse medical NLU tasks. Our ablation experiments show that instruction-tuning on a wider variety of tasks, even when the total number of training instances remains constant, enhances downstream zero-shot generalization.  \footnote{The code and model for this manuscript is available on our project GitHub, \url{https://github.com/uw-bionlp/BioMistral-NLU}.} 

\section*{Introduction}

Fine-tuning large language models (LLMs) on a diverse collection of instruction-following datasets enables LLMs to generalize across a wide range of new tasks in a zero- or few-shot setting.  \cite{chung2022scaling,touvron2023llama} 
Following this instruction fine-tuning phase, medical foundation LLMs  \cite{saab2024capabilities} 
have demonstrated great performance in various medical tasks, which require in-depth medical domain knowledge and logical reasoning ability \cite{nori2023capabilities}, such as medical exams \cite{nori2023capabilities}, common sense reasoning \cite{labrak2024biomistral, han2023medalpaca} and diagnostic reasoning.\cite{saab2024capabilities} This generalizability is particularly crucial for tasks with limited annotated data, where fine-tuning is infeasible. Meanwhile, using a safe, generalized model can also mitigate the safety risks associated with training task-specific models using sensitive medical data.  

\begin{wrapfigure}{r}{0.5\textwidth}
    \vspace{-7pt}
  \begin{center}
    \includegraphics[width=0.45\textwidth]{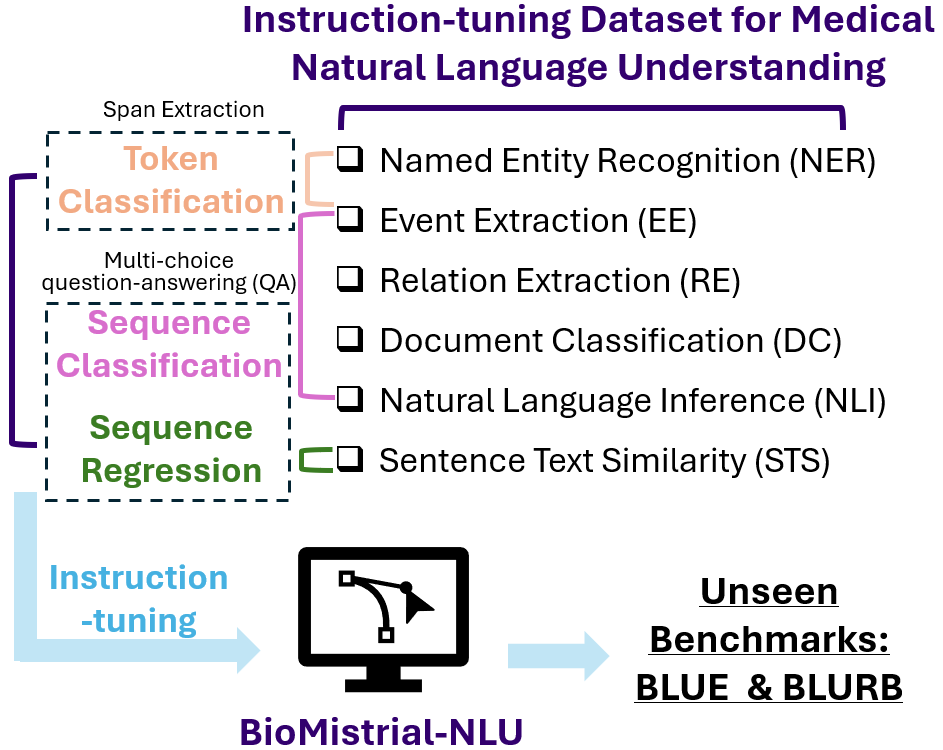}
  \end{center}
  \caption{Instruction-tuning dataset (MNLU-Instruct), system development, and downstream evaluation  for BioMistral-NLU.}
\end{wrapfigure}

Despite their superior generalizability in some areas, instruction-tuned LLMs can underperform smaller-scale, fine-tuned language models, in some specialized medical natural language understanding (NLU) tasks. These tasks require models to understand, interpret, and respond to human language meaningfully.\cite{wang2018glue} Examples of medical NLU tasks include information extraction \cite{xie2024me, hu2023zero} and sentence classification \cite{chen2024evaluating}, which generates additional labels for medical text, facilitating real-time information retrieval and secondary medical research \cite{ajmal2023natural}.
The performance gap may be due to the current foundation LLMs' instruction-tuning phase which focuses primarily on natural language generation (NLG) tasks that allow for free-text, unconstrained outputs.\cite{chung2022scaling} Although many NLG tasks require complex logical reasoning, these skills do not directly translate to nuanced NLU tasks. 

To bridge this gap, we propose a unified prompting format for 7 widely studied medical NLU tasks identified in a recent survey \cite{wang2023pre}, employing span extraction and multi-choice question-answering (QA). Utilizing this unified format, we create an instruction-tuning dataset, MNLU-Instruct, from diverse existing open-source medical NLU corpora. We fine-tune a high-performing biomedical LLM, BioMistral  \cite{labrak2024biomistral} on MNLU-Instruct, resulting in a new, generalizable medical NLU model we call BioMistral-NLU. 
We evaluate the generalizability of BioMistral-NLU, using zero-shot, dataset-agnostic prompts, on two widely adopted benchmark datasets: the Biomedical Language Understanding Evaluation (BLUE)  \cite{peng2019transfer}  and the Biomedical Language Understanding and Reasoning Benchmark (BLURB) \cite{gu2021domain}.Collectively, the benchmarks include 15 biomedical datasets with 6 important NLU task categories, across both clinical and biomedical domains.
In our evaluation, BioMistral-NLU outperforms the original BioMistral, as well as ChatGPT, and GPT-4 on the macro average across all tasks. Our ablation experiments demonstrate that instruction-tuning on a broader variety of tasks, even with a constant total number of training instances, improves downstream zero-shot generalization.. 


\section*{Related work} 
\subsection*{Medical NLU}\label{sec:related_NLU_works}
Within this broad category of medical NLU, there is extensive research on specific NLU tasks in clinical and biomedical domains, such as Information Extraction (IE) and Document Classification (DC).\cite{wu2020deep} To develop a comprehensive understanding of medical NLU, previous research curates two NLU benchmark datasets: the Biomedical Language Understanding Evaluation (BLUE)  \cite{peng2019transfer}  and the Biomedical Language Understanding and Reasoning Benchmark (BLURB)\cite{gu2021domain}. These two benchmarks encompass multiple important medical NLU tasks and are widely adopted to evaluate various LLMs for their medical NLU capabilities. \cite{feng2024evaluation,wang2023large,chen2023extensive}

Previous studies explore the ability of task-agnostic LLMs to perform medical NLU tasks. For example,  Agrawal et al. (2022) \cite{agrawal2022large} demonstrate LLMs' potential for clinical NLU tasks through few-shot in-context learning (ICL).  Hu et al. (2023) \cite{hu2023zero} evaluate ChatGPT on two clinical NER datasets, representing a subset of NLU tasks.
  Wang et al. (2023) \cite{wang2023large} propose a novel prompting strategy for multiple clinical NLU tasks using proprietary LLMs such as ChatGPT \cite{ChatGPT} and GPT-4 \cite{achiam2023gpt}. 
However, they only evaluate the LLMs on a few samples from each task within the BLUE benchmark. 
Similarly, Chen et al. (2023) \cite{chen2023extensive} and  Feng et al. (2024) \cite{feng2024evaluation} systematically evaluate multiple LLMs using the BLURB benchmark \cite{gu2021domain}. 
Although ChatGPT and GPT-4 outperform other LLMs, they considerably underperform the in-domain fine-tuned systems. This performance gap highlights the need for more generalized systems for medical NLU.

\vspace{-8pt}
\subsection*{Instruction tuning for medical NLU}
Instruction tuning involves fine-tuning a pre-trained LM on a diverse collection of instruction-following tasks and thus enables the LM to understand and follow natural language instructions, and generalize to previously unseen tasks in zero-shot or few-shot settings.\cite{chung2022scaling, ouyang2022training} Instruction-tuning datasets typically encompass a wide range of natural language processing (NLP) tasks presented in an instructional format, including reasoning, question-answering, dialogue, and summarization.\cite{zhang2023instruction}
Utilizing instruction tuning, previous research has developed systems focused on generalizing to a limited subset of NLU tasks in the general domain, such as IE \cite{wang2023instructuie, jiao2023instruct, sainz2023gollie, wang2022deepstruct, lu2022unified} and Named Entity Recognition (NER) \cite{zhou2023universalner,zhao2024novel}. 

Several previous studies aim to adapt instruction-tuning to the medical domain, with a major focus on dialogue-based chatbots, such as ChatDoctor \cite{yunxiang2023chatdoctor} and MedAlpaca \cite{han2023medalpaca}. Other medical foundation LLMs, like MedGemini \cite{saab2024capabilities} and Taiyi \cite{luo2024taiyi}, show potential for diverse NLU tasks but lack comprehensive evaluation. Previous system development has often focused on a limited subset of medical NLU tasks. For example, Luo et al. (2022) \cite{luo2022biotabqa} explore Table QA;  Zhao et al. (2024) \cite{zhao2024novel} focused on NER; Sainz et al. (2023) \cite{sainz2023gollie} focused on IE; Rohanian et al. (2023) \cite{rohanian2023exploring} and Tran et al. (2024) \cite{tran2024bioinstruct} focused on QA, IE, and text generation;  
However, the application of these models to other NLU tasks, such as sentence similarity and natural language inference, has not yet been explored. To the best of our knowledge, there is no comprehensive system development and evaluation across all medical NLU tasks for their generalizability. Therefore, in this work, we aim to bridge this gap by evaluating our proposed system in a zero-shot setting using two widely adopted benchmarks, encompassing 7 important medical NLU tasks. 

\section*{Methods}
In this section, we will introduce the task formulation, and outline the three-step approach to creating our generalized LLM across medical NLU tasks.

\vspace{-8pt}
\subsection*{Task formulation}
We reformulate the NLU problem as text generation tasks. Our learning objective $M$ for the medical NLU system is defined by the function $M: (I, X, T) \rightarrow O$. Specifically, given a user instruction $I$, associated medical text $X$, and NLU task labels $T$, the model $M$ is instructed to output the system output $O$, where $I, X, T, O$ correspond to sequences of tokens.  

We developed a unified prompt format, which simplifies evaluation across diverse NLU task outputs, and potentially facilitates knowledge transfer when the system is fine-tuned for a wider range of NLU tasks.

\subsection*{Unified Medical NLU format} 
Building on prior research outlined in the Related Work section, we develop our unified NLU format that focuses on seven critical NLU tasks. Six of these NLU tasks are directly adapted from the BLUE and BLURB benchmarks, including named entity recognition (NER), document classification (DC), relation extraction (RE), multi-choice question-answering (QA), natural language inference (NLI), and semantic text similarity (STS). We also incorporate event extraction (EE), which is extensively researched in the medical domain.\cite{frisoni2021survey} In EE, each event consists of a trigger and multiple arguments that characterize the event. The event trigger extraction (ETE) and event argument extraction (EAE) can be considered as NER. The event argument classification (EAC) classifies the event argument into a subtype, and can be considered as sequence classification. Table \ref{tab:prompts} demonstrates the input-output format for each medical NLU task.

\begin{table*}[h]
\vspace{15pt}
\small
\resizebox{\textwidth}{!}{%
\begin{tabular}{lll}
\hline
\textbf{Task} & \textbf{Input prompt format} & \textbf{Output prompt format} \\ \hline
\begin{tabular}[c]{@{}l@{}}NER/\\ETE\end{tabular} & \begin{tabular}[c]{@{}l@{}}Extract all relevant medical named entities from the medical text below.\\ Focus on identifying following entities: \{$type_1$\}, \{$type_2$\}, ... . \{\textit{text}\}\end{tabular} & \begin{tabular}[c]{@{}l@{}}\{$type_1$\}:\{$span_1$\} ... \{$span_n$\}\\  ...\end{tabular} \\ \hline 
EAE & \begin{tabular}[c]{@{}l@{}}What is the \{$type$\} attribute of the \{$trigger$\} `\{$span$\}' in the medical \\ text below? \{\textit{text}\}\end{tabular} & \{$trigger$\} - \{$attribute$\}:\{$span_1$\} ... \\ \hline 
EAC & \begin{tabular}[c]{@{}l@{}}What is the \{$type$\} attribute of the \{$trigger$\} `\{$span$\}' in the medical \\ text below? \{\textit{text}\} \{\textit{options}\} \end{tabular} & \{$trigger$\} - \{$attribute$\}:\{$option$\} \\ \hline 
DC & \begin{tabular}[c]{@{}l@{}}Which options best describe cancer hallmark from the medical text\\ below? \{\textit{text}\} \{\textit{options}\} \end{tabular} &  \\ \cline{1-2}
RE & \begin{tabular}[c]{@{}l@{}}What is the relation between the\{$type_1$\} entity `\{$span_1$\}' and  the \\ \{$type_2$\}  entity `\{$span_2$\}' from the medical text below? \{\textit{text}\} \{\textit{options}\} \end{tabular} & \\ \cline{1-2}
QA & \{\textit{question}\} \{\textit{text}\} \{\textit{options}\} &  \{$option$\} \\  \cline{1-2}
NLI & \begin{tabular}[c]{@{}l@{}}What is the relation between the premise and hypothesis?\\ Premise: \{$premise$\}. Hypothesis: \{$hypothesis$\} \{\textit{options}\}\end{tabular} & \\ \cline{1-2}
STS & \begin{tabular}[c]{@{}l@{}}How similar are the two sentences below?\\ Sentence 1: \{$sentence_1$\}. Sentence 2: \{$sentence_2$\}. \{\textit{options}\}\end{tabular} &  \\ \hline
\end{tabular}
}
\setlength{\belowcaptionskip}{-5pt}
\caption{\small The task-agnostic prompt format for 7 medical NLU tasks:  named entity recognition (NER), event extraction (EE), document classification (DC), relation extraction (RE), multi-choice question-answering (QA), natural language inference (NLI), and semantic text similarity (STS). Event trigger extraction (ETE), event argument extraction (EAE), and event argument classification (EAC) are all components of the EE task.
\textit{Variables} inside \{\} are derived from each dataset instance. }
\label{tab:prompts}
\vspace{15pt}
\end{table*}

Those seven NLU tasks can be summarized into three categories: (1) token classification, (2) sequence classification, and (3) sequence regression.

NER, ETE, and EAE are \textbf{token classification tasks}, which assign a class label to each token in the input sequence \footnote{ Tasks such as NER are often treated as sequence labeling tasks in the NLP field.\cite{he2020survey} In this work, we refer to them as Token classification tasks for consistency with the BLURB.\cite{gu2021domain}} In token classification, the input includes the user instruction $I$ with pre-defined token labels, and the target text $T$. In the output $O$, each line includes all the token annotations associated with a specific label. Each line starts with a class label, followed by the corresponding positive tokens in the order they appear in $X$. Continuous positive tokens are grouped into text spans (entities), separated by ``...'', for example, ``Disease: fever...headache''. If no tokens are classified as entities, the $O$ is ``None''. More specifically, NER classifies each token as a possible named entity. 

EAC, DC, RE, QA, and NLI are \textbf{sequence classification tasks}, which assign a class label to the entire input token sequence (see Table \ref{tab:other_tasks}). In sequence classification, the user instruction $I$ specifies pre-defined class labels as multiple choices, which is a commonly adopted format in instruction-tuning, for example, ``(B) fevers \textit{happens with} headache''.\cite{chung2022scaling} The system output $O$ is always one or more multi-choice options. In DC, the medical text $X$ is the document. In RE, $X$ is the corresponding medical text snippet with labeled named entities. In NLI, $X$ is a pair of a premise and a hypothesis. In QA, user instruction $I$ involves the task question, and $X$ is the corresponding medical text.

STS is a \textbf{sequence regression task}, which assigns a numeric score to the entire input. In this study, we explore the widely researched task of sequence regression: calculating the semantic text similarity (STS) score between two sentences. 
In the user instruction $I$ of STS, the STS scores correspond to the scoring criteria from the original publication, and are presented as multi-choice options, for example, `(A) The two sentences are on different topics (score 0).'

\begin{multicols}{2}
\begin{table}[H]

\small
\resizebox{0.5\textwidth}{!}{%
\begin{tabular}{ll}
\hline
\textbf{Task} & \textbf{Datasets used for instruction-tuning} \\ \hline
\multirow{2}{*}{NER} & \begin{tabular}[c]{@{}l@{}}i2b2 2006DeID  \cite{uzuner2007evaluating}, i2b2 2011Coreference  \cite{uzuner2012evaluating},\\ i2b2 2012Temporal  \cite{sun2013evaluating}, i2b2 2014 DeID  \cite{stubbs2015annotating}, \\ GENIA \cite{yu2020named}, linnaeus \cite{kocaman2021biomedical}, tmVar \cite{wei2018tmvar}, \\ DrugProt \cite{miranda2023overview}, BioRed \cite{luo2022biored},\\ GNorm \cite{morgan2008overview}, NLM-Gene \cite{islamaj2021nlm}, \\ ClinicalIE \cite{agrawal2022large}, BC4CHEMD \cite{krallinger2015chemdner},\end{tabular} \\ 
 & PubMed PICO  \cite{jin2018pico}, PICO-Data  \cite{nguyen2017aggregating} \\ \hline
EE & \begin{tabular}[c]{@{}l@{}}i2b2 2009Medication \cite{uzuner2010extracting}, i2b2 2018ADE  \cite{henry20202018}, \\ n2c2 2022SDoH \cite{lybarger20232022},\end{tabular} \\ \hline
DC & \begin{tabular}[c]{@{}l@{}}i2b2 2006Smoking \cite{uzuner2008identifying}, i2b2 2008Obesity \cite{uzuner2009recognizing}, \\ n2c2 2018  \cite{stubbs2019cohort}, 2024 SemEval Task 2  \cite{jullien2024semeval}, \\ TrialStop \cite{razuvayevskaya2023clinical}, MTSamples  \cite{MTSample}\end{tabular} \\ \hline
RE & \begin{tabular}[c]{@{}l@{}}i2b2 2011Coreference  \cite{uzuner2012evaluating}, i2b2 2012Temporal  \cite{sun2013evaluating}, \\ EUADR \cite{VANMULLIGEN2012879}, DrugProt \cite{miranda2023overview}, \\ BioRed \cite{luo2022biored}\end{tabular} \\ \hline
NLI & BioNLI  \cite{bastan2022bionli}, SNLI  \cite{bowman2015large}, Multi-NLI \cite{N18-1101} \\ \hline
STS & SIS-B  \cite{wang2018glue} \\ \hline
Summ & PubMedSum  \cite{cohan-etal-2018-discourse}, CDSR  \cite{guo2021automated}, AciDemo \cite{yim2023aci} \\ \hline
\end{tabular}
}
\setlength{\belowcaptionskip}{-5pt}
\caption{The MNLU-Instruct dataset, which is used for fine-tuning: NLU and summarization datasets and tasks curated from existing open-source medical corpora.}
\label{tab:instruction-tuing}
\end{table}

\columnbreak
\begin{table}[H]
\centering
\small
\resizebox{0.45\textwidth}{!}{%
\begin{tabular}{ll}
\hline
\textbf{Dataset} & \textbf{Named entities} \\ \hline
BC2GM & Gene \\ \hline
BC5-chemical & Chemical \\ \hline
BC5-disease & Disease \\ \hline
NCBI-disease & Disease \\ \hline
JNLPBA & \begin{tabular}[c]{@{}l@{}}Protein, Cell type, RNA, Cell line, \\ DNA\end{tabular} \\ \hline
EBM PICO & \begin{tabular}[c]{@{}l@{}}Interventions, Participants, Outcomes \\  \end{tabular} \\ \hline
\end{tabular}
}
\setlength{\belowcaptionskip}{-5pt}
\caption{NER datasets used in the evaluation.}
\label{tab:ner}
\end{table}

\vspace{15pt}
\begin{table}[H]
\small
\centering
\resizebox{0.45\textwidth}{!}{%
\begin{tabular}{lll}
\hline
\textbf{Task} & \textbf{Dataset} & \textbf{Multi-choice options} \\ \hline
DC & HoC & 10 cancer hallmarks \\ 
\hline
\multirow{2}{*}{QA} & PubMedQA & yes / maybe / no \\
 & BioASQ &  yes / no \\ \hline
\multirow{4}{*}{RE} & GAD & 2 gene-disease relations \\  
 & DDI & 4 drug-drug interactions \\  
 & ChemProt & 5 chemical-protein relations \\  
 & i2b2-2010 & 8 medical problem relations \\ 
 \hline
NLI & MedNLI & entails / neutral / contradicts \\ 
\hline
STS & BioSSES & 5 similarity score definitions \\ \hline
\end{tabular}
}
\setlength{\belowcaptionskip}{-5pt}
\caption{Sequence classification and regression datasets used in the evaluation.}
\label{tab:other_tasks}
\end{table}

\end{multicols}

\vspace{-8pt}
\subsection*{MNLU-Instruct dataset}
Focusing on the 7 medical NLU tasks outlined in Table \ref{tab:prompts}, we construct the instruction-tuning dataset, MNLU-Instruct, through intensively searching for publicly available clinical and biomedical NLU datasets outside of BLUE and BLURB. To better assess the generalizability of our proposed system, we intentionally avoid adding any QA datasets to the MNLU-Instruct dataset, using QA tasks as novel tasks specifically for assessment purposes. Instead, beyond NLU tasks, we additionally incorporate three medical summarization tasks, which require similar text summarization and understanding abilities as the QA tasks. Meanwhile, given the limited availability of public medical datasets for NLI and STS, we incorporate datasets from the general domain, including SNLI, Multi-NLI, and SIS-B. As a result, we derive the MNLU-Instruct dataset with the train splits from 33 publicly available datasets shown in Table \ref{tab:instruction-tuing}.  

We construct the NLU input-output pairs in MNLU-Instruct through the task-agnostic prompting strategy shown in Table \ref{tab:prompts}, which directly adapts pre-defined label names from the original publications. We additionally expand abbreviated label names, i.e., from `GENERIF' to `Gene reference into a function (function of a gene)'. 
To increase the variability of MNLU-Instruct, for every NLU input-output pair, we randomly shuffle the order of task labels. Specifically, token labels in token classification tasks, as well as multi-choice options in sequence classification and regression tasks, are randomly shuffled. When train splits are unavailable or datasets have very few input-output pairs, we utilize the entire datasets for training. The complete set of dataset labels, prompts, and statistics can be found in our project GitHub Repository (\url{https://github.com/uw-bionlp/BioMistral-NLU}.)


\vspace{-8pt}
\subsection*{BioMistral-NLU system development}
We hypothesize that instruction-tuning on a diverse, yet relevant set of tasks improves the generalizability of LLMs on medical NLU tasks. To verify this hypothesis, we fine-tune a high-performing medical LLM on MNLU-Instruct and evaluate it in a zero-shot setting.

We chose BioMistral-7B-DARE as our baseline system, which is the state-of-the-art open-source LLM on multiple medical QA tasks. For simplicity, we refer to BioMistral-7B-DARE as BioMistral in this work. 
We fine-tune BioMistral using all of the parameters on MNLU-Instruct, resulting in BioMistral-NLU-FT. However, fine-tuning LLMs in specialized domains can potentially degrade their original generalization ability across broader tasks. \cite{ainsworth2022git} To mitigate this risk and preserve the versatility of the original BioMistral, we utilize DARE  \cite{yu2023language}, as suggested by Labrak et al. (2024)\cite{labrak2024biomistral}. This approach integrates model parameters from BioMistral-NLU-FT and BioMistral, without additional training, and creates the merged system \textbf{BioMistral-NLU}.

The experiment is conducted using the alignment-handbook\footnote{\url{https://github.com/huggingface/alignment-handbook}} package.  Based on the engineering judgment recommended by the alignment-handbook GitHub discussion, we set the number of epochs to 3, the batch size to 16, and configured the learning rate to 2e-04 with a warm-up ratio of 0.1, using 4 A100 GPUs. The rest hyperparameters are the same as the default configurations by the alignment-handbook. For inference, we use the vllm package\footnote{\url{https://github.com/vllm-project/vllm}} and set the temperature to 0. Our whole fine-tuning process for BioMistral-NLU takes less than one day.

\section*{Experiment setup}
In this section, we will introduce our evaluation datasets, evaluation metrics, and comparative systems.

\subsection*{Evaluation datasets and metrics}
We evaluate BioMistral-NLU in a zero-shot setting using BLURB and BLUE. Due to the sensitivity involved in the deployment of clinical note-based corpora, we excluded two inaccessible datasets from BLUE,  ShARe/CLEF  \cite{suominen2013overview} and MedSTS \cite{wang2020medsts}. Some datasets are included in both benchmarks evaluated, resulting in a total of 7 tasks and 15 unique datasets evaluated. 
We developed the evaluation datasets using the unified prompt format outlined in Table \ref{tab:prompts};
the entity types and multi-choice options for these datasets are shown in Table \ref{tab:ner} and \ref{tab:other_tasks}. 

For consistency with prior studies, we utilize the same evaluation criteria from BLUE  \cite{peng2019transfer} and BLURB \cite{gu2021domain}. Token classification tasks are evaluated using F1 scores at the entity level, except for the PICO dataset, which is evaluated at the token level. When class labels are balanced like in NLI and QA, sequence classification tasks are evaluated using accuracy. When class labels are imbalanced, like in RE, sequence classification tasks are evaluated using F1. For the sequence regression task, STS, system outputs are converted to numerical integer scores and evaluated based on Pearson correlation.

\subsection*{Comparative systems} \label{sec:sec_baseline}
We compare our proposed system, BioMistral-NLU, with our baseline,  BioMistral, as well as other high-performing systems.

\textbf{Proprietary LLMs}: \textbf{ChatGPT} \cite{ChatGPT} and \textbf{GPT-4} \cite{achiam2023gpt}.  We reference prior research that evaluates these proprietary LLMs on BLURB \cite{chen2023extensive, feng2024evaluation} \footnote{GPT-4 version: \href{https://platform.openai.com/docs/models/gpt-4-turbo-and-gpt-4}{gpt-4-0613}. ChatGPT version: GPT-3.5, though the exact version was not specified in the original publication.} Note that ChatGPT's performance is reported under one-shot ICL, while GPT-4's performance is based on randomly selected 3-shot examples for NER tasks and zero-shot for other tasks. Additionally, their prompts are strategically optimized for each dataset, resulting in competitive systems. Given that Feng et al. (2024) \cite{feng2024evaluation} demonstrated GPT-4's superiority over FLAN-T5-XXL \cite{chung2024scaling}, PMC- LLaMA-13B \cite{wu2024pmc}, and Zephyr-7B-Beta \cite{tunstall2023zephyr}, we excluded these systems from further evaluation.

\textbf{Open-source LLMs}: \textbf{BioMistral}\footnote{\url{https://huggingface.co/BioMistral/BioMistral-7B}}, and \textbf{ LLaMA-3.1-8B-Instruct} \footnote{\url{https://huggingface.co/meta-llama/Llama-3.1-8B-Instruct}} \cite{ LLaMA3}. In our controlled experiments, we evaluate open-source LLMs using our proposed unified prompting formats shown in Table \ref{tab:prompts}. The evaluation is conducted in a zero-shot setting, except for NER datasets. Since 
our desired token classification output prompt format is less common during those open-source LLMs' instruction tuning phase, 
we additionally incorporate an explanation for the output formats and two in-context examples. For each inference query, the 2-shot examples are randomly selected from the training split of each dataset. We ensure that the outputs from the 2-shot examples are distinct from each other to prevent bias toward a specific answer. 



\vspace{10pt}
\begin{table}[H]
\small
\centering
\resizebox{0.95\textwidth}{!}{%
\begin{tabular}{llllccccc}
\hline
\multirow{2}{*}{\textbf{Task}} & \multirow{2}{*}{\textbf{\begin{tabular}[c]{@{}l@{}}Evaluation\\ Metric\end{tabular}}} & \multirow{2}{*}{\textbf{Dataset}} & \multirow{2}{*}{\textbf{\begin{tabular}[c]{@{}l@{}}Num\\ Instances\end{tabular}}} & \multirow{2}{*}{\textbf{\begin{tabular}[c]{@{}c@{}}Chat-\\ GPT\cite{chen2023extensive}\end{tabular}}} & \multirow{2}{*}{\textbf{GPT-4\cite{feng2024evaluation}}} & \multirow{2}{*}{\textbf{\begin{tabular}[c]{@{}c@{}}  LLaMA\\ -3.1-8B\end{tabular}}} & \multicolumn{2}{c}{\textbf{BioMistral-7B}} \\ \cline{8-9} 
 &  &  &  &  &  &  & \textbf{Baseline} & \textbf{MNLU} \\ \hline
\multirow{6}{*}{NER} & \multirow{5}{*}{\begin{tabular}[c]{@{}l@{}}Entity-\\ level F1\end{tabular}} & BC2GM$^\dagger$ & \multicolumn{1}{l|}{6,322} & 37.5 & \multicolumn{1}{c|}{54.6} & 34.0 & 34.1 & {\ul \textbf{61.5}} \\
 &  & BC5-chemical$^\dagger$* & \multicolumn{1}{l|}{5,385} & 60.3 & \multicolumn{1}{c|}{78.2} & 46.0 & 45.0 & {\ul \textbf{89.9}} \\
 &  & BC5-disease$^\dagger$* & \multicolumn{1}{l|}{4,424} & 51.8 & \multicolumn{1}{c|}{63.9} & 35.2 & 33.7 & {\ul \textbf{67.0}} \\
 &  & NCBI-disease$^\dagger$ & \multicolumn{1}{l|}{955} & 50.5 & \multicolumn{1}{c|}{66.0} & 49.4 & 39.9 & {\ul \textbf{61.8}} \\
 &  & JNLPBA$^\dagger$ & \multicolumn{1}{l|}{8,657} & 41.3 & \multicolumn{1}{c|}{45.4} & 15.0 & 25.6 & {\ul \textbf{64.4}} \\ \cline{2-3}
 & \begin{tabular}[c]{@{}l@{}}Token-\\ level F1\end{tabular} & EBM PICO$^\dagger$ & \multicolumn{1}{l|}{24,474} & 55.6 & \multicolumn{1}{c|}{33.5} & 12.2 & 19.6 & \textbf{55.3} \\ \cline{1-3}
DC & F1 & HoC$^\dagger$* & \multicolumn{1}{l|}{315} & 51.2 & \multicolumn{1}{c|}{62.5} & 37.6 & 47.3 & {\ul \textbf{63.8}} \\ \cline{1-3}
\multirow{2}{*}{QA} & \multirow{2}{*}{Acc} & PubMedQA$^\dagger$ & \multicolumn{1}{l|}{500} & 76.5 & \multicolumn{1}{c|}{70.6} & 73.4 & 72.0 & 70.2 \\
 &  & BioASQ$^\dagger$ & \multicolumn{1}{l|}{263} & 88.6 & \multicolumn{1}{c|}{85.7} & 92.0 & 74.9 &  86.7 \\ \cline{1-3}
\multirow{4}{*}{RE} & \multirow{4}{*}{F1} & GAD$^\dagger$ & \multicolumn{1}{l|}{534} & 52.4 & \multicolumn{1}{c|}{51.5} & 57.7 & 55.0 & {\ul \textbf{58.5}} \\
 &  & DDI$^\dagger$* & \multicolumn{1}{l|}{5,761} & 51.6 & \multicolumn{1}{c|}{37.7} & 22.2 & 10.0 & 13.0 \\
 &  & ChemProt$^\dagger$* & \multicolumn{1}{l|}{14,744} & 34.2 & \multicolumn{1}{c|}{37.6} & 40.4 & 28.6 & {\ul 38.1} \\
 &  & i2b2-2010* & \multicolumn{1}{l|}{6,292} & - & \multicolumn{1}{c|}{-} & 40.6 & 30.9 & {\ul \textbf{41.8}} \\ \cline{1-3}
NLI & Acc & MedNLI* & \multicolumn{1}{l|}{1,422} & - & \multicolumn{1}{c|}{-} & 44.8 & 49.3 & {\ul \textbf{57.5}} \\ \cline{1-3}
STS & \begin{tabular}[c]{@{}l@{}}Pearson\\ Corr\end{tabular} & BioSSES$^\dagger$* & \multicolumn{1}{l|}{20} & 42.8 & \multicolumn{1}{c|}{89.3} & 68.7 & 69.1 & {\ul \textbf{80.8}} \\ \hline
\multirow{2}{*}{\begin{tabular}[c]{@{}l@{}}Bench-\\ Mark\end{tabular}} & \multirow{2}{*}{\begin{tabular}[c]{@{}l@{}}Macro\\ AVG\end{tabular}} & BLURB$^\dagger$ & \multicolumn{1}{l|}{-} & 53.4 & \multicolumn{1}{c|}{59.7} & 44.9 & 42.7 & {\ul \textbf{62.4}} \\
 &  & BLUE* & \multicolumn{1}{l|}{-} & - & \multicolumn{1}{c|}{-} & 41.9 & 39.2 & {\ul \textbf{56.5}} \\ \hline
\end{tabular}

}
\setlength{\belowcaptionskip}{-5pt}
\caption{Our proposed system, BioMistral-NLU's zero-shot performance on 15 unseen medical NLU datasets from 2 benchmarks: BLURB (labeled by $^\dagger$) and BLUE (labeled by $^*$). \textbf{Bold} indicates superior performance over other open-source LLMs, which utilize the same, dataset-agnostic prompts as BioMistral-NLU, under the zero-shot setting, except for two extra random examples for NER tasks. \underline{Underline} indicates better performance over  ChatGPT one-shot and GPT-4 three-shot ICL. ChatGPT and GPT-4 utilize dataset-specific prompts and few-shot examples, and therefore have advantages over our proposed system. `-' indicates that the performance is not measured by prior research. }
\label{tab:results}
\end{table} 


\section*{Results}

%
Following the practice in BLURB \cite{gu2021domain}, we average system performance across datasets for an overview. As shown in Table \ref{tab:results}, BioMistral-NLU outperforms the baseline BioMistral with an increase in the macro average score of 19.7 for BLURB and 16.7 for BLUE. Meanwhile, BioMistral-NLU outperforms the proprietary models, achieving an increase in the macro average score of 9.0 over ChatGPT, and 2.7 over GPT-4  for BLURB. Our results demonstrate that instruction-tuning on diverse medical NLU tasks using our unified format effectively improves the LLMs' generalizability to unseen NLU datasets. In this section, we will analyze the results and characterize the gaps between the systems.
\vspace{-8pt}
\subsection*{Comparison across systems}
Comparing BioMistral-NLU with the baseline BioMistral, we observe an average performance increase of 33.7 for NER tasks and 8.2 for other tasks. This difference may originate from the instruct-tuning phase of BioMistral. While the number of NER task instances might be relatively small during BioMistral's instruction-tuning phase, the other tasks utilize a QA prompting strategy and are likely similar to some of BioMistral's instruction-tuning tasks. This necessitates instruction-tuning on a wider variety of NLU tasks to improve the LLM's generalizability. 

Comparing BioMistral-NLU with proprietary LLMs in the BLURB benchmark, we observe that BioMistral-NLU has an average F1 score of 9.7 higher than GPT-4 across NER tasks. However, for other BLURB tasks, BioMistral-NLU has an average score of 2.0 higher than ChatGPT and  5.4 lower than GPT-4. Given that GPT-4 is significantly larger in terms of parameter size and has been instruction-tuned on much more diverse corpora, its superior generalization ability for other tasks involving more complex reasoning is consistent with the empirical scaling law. \cite{kaplan2020scaling,chung2022scaling}




\vspace{-8pt}
\subsection*{Error analysis}

We observe that for NER tasks, a major source of error for BioMistral-NLU is the nuanced task of accurately identifying exact named entity boundaries. For example, in the BC2GM gene NER dataset, the predicted named entity is `Id - 1', whereas the gold named entity is `mouse Id - 1'. To better understand the prevalence of this discrepancy, we evaluate 5 NER datasets using a relaxed criterion, where two named entities are considered equivalent if their spans overlap. 
Using this relaxed criterion, we observe an average improvement of 15.5 in F1 across the 5 NER datasets from the original entity-level F1.


In all RE tasks, BioMistral-NLU demonstrates recall rates that are 10 to 70 points higher than its precision, suggesting a tendency to identify many false positive relationships. One major source of these false positives is the occurrence of interactions between entities, which do not fit into any of the pre-defined relation categories of interest. As a result, BioMistral-NLU assigns a wrong relation label instead of recognizing no relation.

In the sequence regression dataset, BioSSES, BioMistral-NLU tends to predict intermediate similarity scores (such as scores of 2 or 3) rather than extreme scores (0, 1, 4, or 5).
\section*{Discussion} 
In this section, we will evaluate the impact of instruction dataset composition, focusing on two components: instruction-tuning tasks and domains.

\vspace{-8pt}
\subsection*{Impact of instruction-tuning tasks} \label{sec:abl_tasks}
We aim to assess the impact of instruction-tuning task selection from two perspectives: (1) its relevance to downstream tasks and (2) its task diversity. Focusing on these two perspectives, we fine-tune the baseline system, BioMistral, with different subsets of tasks used to build BioMistral-NLU. We evaluate the fine-tuned system on the 4 RE datasets from Table \ref{tab:results} in a zero-shot setting, and compare the macro-average F1 scores across the 4 RE datasets. 

To study the impact of task relevance, we first construct two instruction-tuning setups: (1) with the RE task (\textbf{w/ RE}) and (2) with the DC task (\textbf{w/o RE}). We chose the DC task because DC employs a similar QA prompting format to RE and it contains 6 diverse datasets from Table \ref{tab:instruction-tuing}. To study the impact of task diversity, besides DC and RE, we additionally include 2 and 4 more randomly selected tasks from Table \ref{tab:instruction-tuing}. 

All fine-tuning experiments are controlled using a fixed number of 50,000 data instances and running for three epochs. We maintain an equal number of instances for each task (i.e., 50,000/k instances per task when fine-tuning with k tasks), and randomly sample fine-tuning instances from all datasets within the same task. 

\begin{multicols}{2}
More specifically, our experiment settings are:

\begin{enumerate}[nolistsep]
    \item \textbf{w/ RE}:\begin{enumerate}[nolistsep]
        \item 1 task: RE
        \item 3 tasks: RE, NLI, NER
        \item 5 tasks: RE, NLI, NER, EE, STS
    \end{enumerate}
    
    \item \textbf{w/o RE}:\begin{enumerate}[nolistsep]
        \item 1 task: DC
        \item 3 tasks: DC, NLI, NER
        \item 5 tasks: DC, NLI, NER, EE, STS
    \end{enumerate}
\end{enumerate}
\columnbreak
\begin{figure}[H]
\centering
\includegraphics[width=0.5\textwidth]{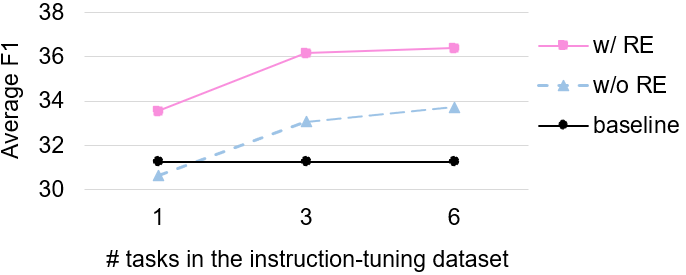}
\vspace{2pt}
\caption{Average zero-shot performance on the 4 RE datasets, after instruction-tuning on 50k instances.} \label{fig:abalation_task} 

\end{figure}
\end{multicols}

After BioMistral is fine-tuned with the same number of instances, we observe the following from Figure \ref{fig:abalation_task}: (1) Overall, setting 1 (with RE) consistently outperforms setting 2 (without RE), due to its relevance to the RE datasets used in downstream evaluation; (2) In both settings, system performance increases with the number of fine-tuning tasks, demonstrating the benefits of fine-tuning with multiple tasks; (3) When fine-tuning on a single task, the performance improvement on downstream tasks depends on the similarity between fine-tuning task and the downstream task.




\subsection*{Impact of instruction-tuning domain}
After demonstrating the benefits of diverse instruction-tuning tasks, we now examine individual tasks. Note that the BLUE benchmark includes both biomedical and clinical datasets: biomedical data is derived from scientific publications, while clinical data consists of semi-structured clinical notes from patients.\cite{wu2011semantic} In this section, we assess how domain selection affects downstream generalizability.

\begin{wrapfigure}{r}{0.5\textwidth}
\vspace{7pt}
  \begin{center}
    \includegraphics[width=0.45\textwidth]{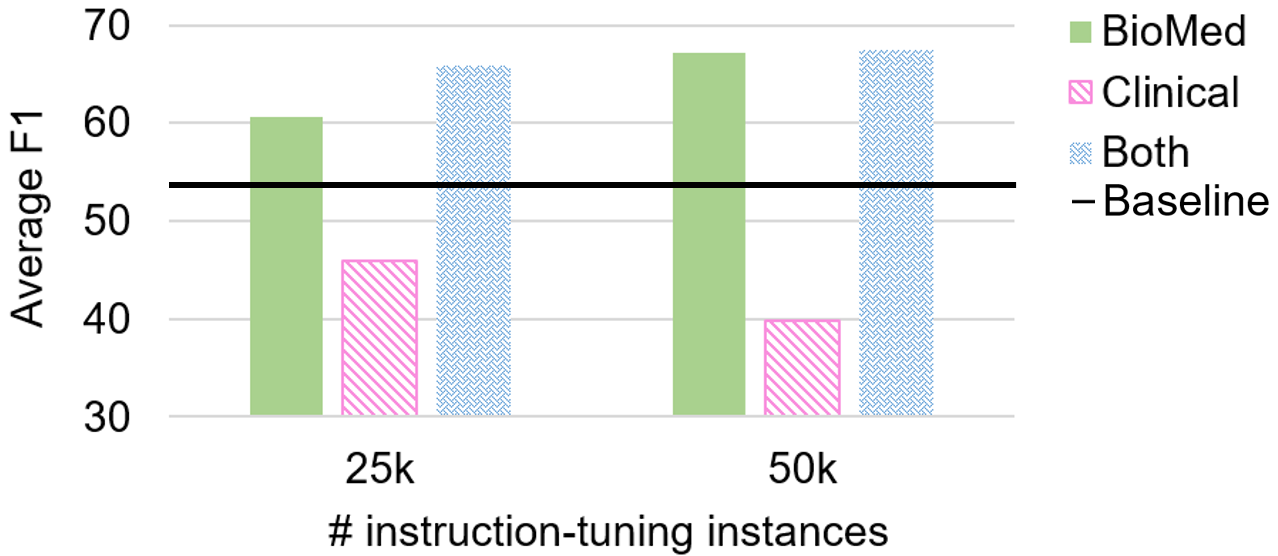}
    \caption{Average zero-shot performance on 6 biomedical NER datasets, when finetuned on different domains.} \label{fig:abalation_domain} 
  \end{center}

\vspace{6pt}
\end{wrapfigure}

We follow a similar experimental setup as described in the previous section,  
fine-tuning BioMistral for three epochs over 25,000 data instances. The fine-tuned system is evaluated on six biomedical NER datasets from Table \ref{tab:results} in a zero-shot setting, using macro average F1 scores. Instruction-tuning NER datasets from MNLU-Instruct \footnote{We also include event triggers as named entities.} are divided into biomedical and clinical splits. Our experiments include fine-tuning on a single split (\textbf{BioMed} / \textbf{Clinical}) and both splits (\textbf{Both}). We additionally combine single splits or include additional instances, creating a similar experiment setting with 50k instances. We use the 2-shot BioMistral as the \textbf{baseline system}. 


From Figure \ref{fig:abalation_domain}, we observe the following: (1) Instruction-tuning on the BioMed domain alone consistently outperforms tuning on the Clinical domain alone when using the same number of instances. (2) Compared to the baseline, instruction-tuning on the Clinical domain negatively impacts downstream performance on the BioMed domain. (3) Combining instances from both domains improves downstream generalizability to the BioMed domain, even with the same total number of instances. (4) Increasing the number of instances from the BioMed or Both domains improves performance, while adding more instances solely from the Clinical domain decreases performance.

\section*{Conclusion}
In this work, we introduce a unified prompting format for 7 important medical NLU tasks, and develop an instruction-tuning dataset based on publicly available clinical and biomedical corpora. Our experiment demonstrates that fine-tuning across diverse medical NLU datasets improves the system's generalizability in a zero-shot setting with dataset-agnostic prompt tuning. 
Our ablation study 
underscores the necessity for instruction tuning across diverse medical NLU tasks, including domain-specific lexicon and common biomedical tasks.

Our NLU-focused, instruction-tuning pipeline could be applied to other LLMs beyond BioMistral. We conducted an initial fine-tuning experiment with the  LLaMA-3.1-8B model, using the same experiment configuration and MNLU-Instruct dataset as in BioMistral-NLU. While there were improvements for some evaluation datasets, the overall performance gains were much smaller compared to the BioMistral-NLU. In the future, we aim to better understand the optimal instruction-tuning configurations for different LLMs.

Our future work will also focus on further improving the generalized LLM's zero-shot performance on medical NLU tasks and narrowing its gap to in-domain fine-tuned systems. 
Since LLMs are often known to struggle with adhering to in-context annotation guidelines \cite{zhang2023promptner}, our future work will focus on integrating nuanced task descriptions from annotation guidelines into both the fine-tuning and inference stages.\cite{sainz2023gollie} Future work could also involve a self-verification step  \cite{gero2023self} or using a knowledge base as augmentation  \cite{lewis2020retrieval} to reduce false positives in the sequence classification tasks.

\section*{Acknowledgement}
This work was supported by the National Institutes of Health (NIH)—National Cancer Institute (Grant Nos. 1R01CA248422-01A1), National Library of Medicine (Grant No. 2R15LM01320902A1), and National Center for Advancing Translational Sciences of the National Institutes of Health (Grant No. UL1TR002319).  The content is solely the responsibility of the authors and does not necessarily represent the official views of the NIH.

\small
\bibliography{custom}

\begin{thebibliography}{10}

\bibitem{chung2022scaling}
Chung HW, Hou L, Longpre S, Zoph B, Tay Y, Fedus W, et~al.
\newblock Scaling Instruction-Finetuned Language Models.
\newblock arXiv e-prints. 2022:arXiv-2210.

\bibitem{touvron2023llama}
Touvron H, Martin L, Stone K, Albert P, Almahairi A, Babaei Y, et~al.
\newblock Llama 2: Open foundation and fine-tuned chat models.
\newblock arXiv preprint arXiv:230709288. 2023.

\bibitem{saab2024capabilities}
Saab K, Tu T, Weng WH, Tanno R, Stutz D, Wulczyn E, et~al.. Capabilities of Gemini Models in Medicine; 2024.

\bibitem{nori2023capabilities}
Nori H, King N, McKinney SM, Carignan D, Horvitz E.
\newblock Capabilities of gpt-4 on medical challenge problems.
\newblock arXiv preprint arXiv:230313375. 2023.

\bibitem{labrak2024biomistral}
Labrak Y, Bazoge A, Morin E, Gourraud PA, Rouvier M, Dufour R.
\newblock BioMistral: A Collection of Open-Source Pretrained Large Language Models for Medical Domains.
\newblock arXiv preprint arXiv:240210373. 2024.

\bibitem{han2023medalpaca}
Han T, Adams LC, Papaioannou JM, Grundmann P, Oberhauser T, L{\"o}ser A, et~al.
\newblock MedAlpaca--an open-source collection of medical conversational AI models and training data.
\newblock arXiv preprint arXiv:230408247. 2023.

\bibitem{wang2018glue}
Wang A, Singh A, Michael J, Hill F, Levy O, Bowman SR.
\newblock GLUE: A multi-task benchmark and analysis platform for natural language understanding.
\newblock arXiv preprint arXiv:180407461. 2018.

\bibitem{xie2024me}
Xie Q, Chen Q, Chen A, Peng C, Hu Y, Lin F, et~al.
\newblock Me LLaMA: Foundation Large Language Models for Medical Applications.
\newblock arXiv preprint arXiv:240212749. 2024.

\bibitem{hu2023zero}
Hu Y, Ameer I, Zuo X, Peng X, Zhou Y, Li Z, et~al.
\newblock Zero-shot clinical entity recognition using chatgpt.
\newblock arXiv preprint arXiv:230316416. 2023.

\bibitem{chen2024evaluating}
Chen S, Li Y, Lu S, Van H, Aerts HJ, Savova GK, et~al.
\newblock Evaluating the ChatGPT family of models for biomedical reasoning and classification.
\newblock Journal of the American Medical Informatics Association. 2024;31(4):940-8.

\bibitem{ajmal2023natural}
Ajmal S, Ahmed AAI, Jalota C.
\newblock Natural language processing in improving information retrieval and knowledge discovery in healthcare conversational agents.
\newblock Journal of Artificial Intelligence and Machine Learning in Management. 2023;7(1):34-47.

\bibitem{wang2023pre}
Wang B, Xie Q, Pei J, Chen Z, Tiwari P, Li Z, et~al.
\newblock Pre-trained language models in biomedical domain: A systematic survey.
\newblock ACM Computing Surveys. 2023;56(3):1-52.

\bibitem{peng2019transfer}
Peng Y, Yan S, Lu Z.
\newblock Transfer Learning in Biomedical Natural Language Processing: An Evaluation of BERT and ELMo on Ten Benchmarking Datasets.
\newblock BioNLP 2019. 2019:58.

\bibitem{gu2021domain}
Gu Y, Tinn R, Cheng H, Lucas M, Usuyama N, Liu X, et~al.
\newblock Domain-specific language model pretraining for biomedical natural language processing.
\newblock ACM Transactions on Computing for Healthcare (HEALTH). 2021;3(1):1-23.

\bibitem{wu2020deep}
Wu S, Roberts K, Datta S, Du J, Ji Z, Si Y, et~al.
\newblock Deep learning in clinical natural language processing: a methodical review.
\newblock Journal of the American Medical Informatics Association. 2020;27(3):457-70.

\bibitem{feng2024evaluation}
Feng H, Ronzano F, LaFleur J, Garber M, de~Oliveira R, Rough K, et~al.
\newblock Evaluation of large language model performance on the Biomedical Language Understanding and Reasoning Benchmark.
\newblock medRxiv. 2024:2024-05.

\bibitem{wang2023large}
Wang Y, Zhao Y, Petzold L.
\newblock Are large language models ready for healthcare? a comparative study on clinical language understanding.
\newblock In: Machine Learning for Healthcare Conference. PMLR; 2023. p. 804-23.

\bibitem{chen2023extensive}
Chen Q, Sun H, Liu H, Jiang Y, Ran T, Jin X, et~al.
\newblock An extensive benchmark study on biomedical text generation and mining with ChatGPT.
\newblock Bioinformatics. 2023;39(9):btad557.

\bibitem{agrawal2022large}
Agrawal M, Hegselmann S, Lang H, Kim Y, Sontag D.
\newblock Large Language Models are Few-Shot Clinical Information Extractors.
\newblock In: Proceedings of the 2022 Conference on Empirical Methods in Natural Language Processing; 2022. .

\bibitem{ChatGPT}
OpenAI: Introducing ChatGPT; 2022.
\newblock Accessed: 2024-04-12.
\newblock \url{https://openai.com/blog/chatgpt}.

\bibitem{achiam2023gpt}
Achiam J, Adler S, Agarwal S, Ahmad L, Akkaya I, Aleman FL, et~al.
\newblock Gpt-4 technical report.
\newblock arXiv preprint arXiv:230308774. 2023.

\bibitem{ouyang2022training}
Ouyang L, Wu J, Jiang X, Almeida D, Wainwright C, Mishkin P, et~al.
\newblock Training language models to follow instructions with human feedback.
\newblock Advances in neural information processing systems. 2022;35:27730-44.

\bibitem{zhang2023instruction}
Zhang S, Dong L, Li X, Zhang S, Sun X, Wang S, et~al.
\newblock Instruction tuning for large language models: A survey.
\newblock arXiv preprint arXiv:230810792. 2023.

\bibitem{wang2023instructuie}
Wang X, Zhou W, Zu C, Xia H, Chen T, Zhang Y, et~al.
\newblock InstructUIE: multi-task instruction tuning for unified information extraction.
\newblock arXiv preprint arXiv:230408085. 2023.

\bibitem{jiao2023instruct}
Jiao Y, Zhong M, Li S, Zhao R, Ouyang S, Ji H, et~al.
\newblock Instruct and extract: Instruction tuning for on-demand information extraction.
\newblock arXiv preprint arXiv:231016040. 2023.

\bibitem{sainz2023gollie}
Sainz O, Garc{\'\i}a-Ferrero I, Agerri R, de~Lacalle OL, Rigau G, Agirre E.
\newblock Gollie: Annotation guidelines improve zero-shot information-extraction.
\newblock arXiv preprint arXiv:231003668. 2023.

\bibitem{wang2022deepstruct}
Wang C, Liu X, Chen Z, Hong H, Tang J, Song D.
\newblock DeepStruct: Pretraining of language models for structure prediction.
\newblock arXiv preprint arXiv:220510475. 2022.

\bibitem{lu2022unified}
Lu Y, Liu Q, Dai D, Xiao X, Lin H, Han X, et~al.
\newblock Unified structure generation for universal information extraction.
\newblock arXiv preprint arXiv:220312277. 2022.

\bibitem{zhou2023universalner}
Zhou W, Zhang S, Gu Y, Chen M, Poon H.
\newblock Universalner: Targeted distillation from large language models for open named entity recognition.
\newblock arXiv preprint arXiv:230803279. 2023.

\bibitem{zhao2024novel}
Zhao J, Liu C, Liang J, Li Z, Xiao Y.
\newblock A Novel Cascade Instruction Tuning Method for Biomedical NER.
\newblock In: ICASSP 2024-2024 IEEE International Conference on Acoustics, Speech and Signal Processing (ICASSP). IEEE; 2024. p. 11701-5.

\bibitem{yunxiang2023chatdoctor}
Yunxiang L, Zihan L, Kai Z, Ruilong D, You Z.
\newblock Chatdoctor: A medical chat model fine-tuned on llama model using medical domain knowledge.
\newblock arXiv preprint arXiv:230314070. 2023.

\bibitem{luo2024taiyi}
Luo L, Ning J, Zhao Y, Wang Z, Ding Z, Chen P, et~al.
\newblock Taiyi: a bilingual fine-tuned large language model for diverse biomedical tasks.
\newblock Journal of the American Medical Informatics Association. 2024:ocae037.

\bibitem{luo2022biotabqa}
Luo M, Saxena S, Mishra S, Parmar M, Baral C.
\newblock Biotabqa: Instruction learning for biomedical table question answering.
\newblock arXiv preprint arXiv:220702419. 2022.

\bibitem{rohanian2023exploring}
Rohanian O, Nouriborji M, Clifton DA.
\newblock Exploring the Effectiveness of Instruction Tuning in Biomedical Language Processing.
\newblock arXiv preprint arXiv:240100579. 2023.

\bibitem{tran2024bioinstruct}
Tran H, Yang Z, Yao Z, Yu H.
\newblock BioInstruct: instruction tuning of large language models for biomedical natural language processing.
\newblock Journal of the American Medical Informatics Association. 2024:ocae122.

\bibitem{frisoni2021survey}
Frisoni G, Moro G, Carbonaro A.
\newblock A survey on event extraction for natural language understanding: Riding the biomedical literature wave.
\newblock IEEE Access. 2021;9:160721-57.

\bibitem{he2020survey}
He Z, Wang Z, Wei W, Feng S, Mao X, Jiang S.
\newblock A survey on recent advances in sequence labeling from deep learning models.
\newblock arXiv preprint arXiv:201106727. 2020.

\bibitem{uzuner2007evaluating}
Uzuner {\"O}, Luo Y, Szolovits P.
\newblock Evaluating the state-of-the-art in automatic de-identification.
\newblock Journal of the American Medical Informatics Association. 2007;14(5):550-63.

\bibitem{uzuner2012evaluating}
Uzuner O, Bodnari A, Shen S, Forbush T, Pestian J, South BR.
\newblock Evaluating the state of the art in coreference resolution for electronic medical records.
\newblock Journal of the American Medical Informatics Association. 2012;19(5):786-91.

\bibitem{sun2013evaluating}
Sun W, Rumshisky A, Uzuner O.
\newblock Evaluating temporal relations in clinical text: 2012 i2b2 challenge.
\newblock Journal of the American Medical Informatics Association. 2013;20(5):806-13.

\bibitem{stubbs2015annotating}
Stubbs A, Uzuner {\"O}.
\newblock Annotating longitudinal clinical narratives for de-identification: The 2014 i2b2/UTHealth corpus.
\newblock Journal of biomedical informatics. 2015;58:S20-9.

\bibitem{yu2020named}
Yu J, Bohnet B, Poesio M.
\newblock Named entity recognition as dependency parsing.
\newblock arXiv preprint arXiv:200507150. 2020.

\bibitem{kocaman2021biomedical}
Kocaman V, Talby D.
\newblock Biomedical named entity recognition at scale.
\newblock In: Pattern Recognition. ICPR International Workshops and Challenges: Virtual Event, January 10--15, 2021, Proceedings, Part I. Springer; 2021. p. 635-46.

\bibitem{wei2018tmvar}
Wei CH, Phan L, Feltz J, Maiti R, Hefferon T, Lu Z.
\newblock tmVar 2.0: integrating genomic variant information from literature with dbSNP and ClinVar for precision medicine.
\newblock Bioinformatics. 2018;34(1):80-7.

\bibitem{miranda2023overview}
Miranda-Escalada A, Mehryary F, Luoma J, Estrada-Zavala D, Gasco L, Pyysalo S, et~al.
\newblock Overview of DrugProt task at BioCreative VII: data and methods for large-scale text mining and knowledge graph generation of heterogenous chemical--protein relations.
\newblock Database. 2023;2023:baad080.

\bibitem{luo2022biored}
Luo L, Lai PT, Wei CH, Arighi CN, Lu Z.
\newblock BioRED: a rich biomedical relation extraction dataset.
\newblock Briefings in Bioinformatics. 2022;23(5):bbac282.

\bibitem{morgan2008overview}
Morgan AA, Lu Z, Wang X, Cohen AM, Fluck J, Ruch P, et~al.
\newblock Overview of BioCreative II gene normalization.
\newblock Genome biology. 2008;9:1-19.

\bibitem{islamaj2021nlm}
Islamaj R, Wei CH, Cissel D, Miliaras N, Printseva O, Rodionov O, et~al.
\newblock NLM-Gene, a richly annotated gold standard dataset for gene entities that addresses ambiguity and multi-species gene recognition.
\newblock Journal of biomedical informatics. 2021;118:103779.

\bibitem{krallinger2015chemdner}
Krallinger M, Rabal O, Leitner F, Vazquez M, Salgado D, Lu Z, et~al.
\newblock The CHEMDNER corpus of chemicals and drugs and its annotation principles.
\newblock Journal of cheminformatics. 2015;7:1-17.

\bibitem{jin2018pico}
Jin D, Szolovits P.
\newblock Pico element detection in medical text via long short-term memory neural networks.
\newblock In: Proceedings of the BioNLP 2018 workshop; 2018. p. 67-75.

\bibitem{nguyen2017aggregating}
Nguyen AT, Wallace BC, Li JJ, Nenkova A, Lease M.
\newblock Aggregating and predicting sequence labels from crowd annotations.
\newblock In: Proceedings of the conference. Association for Computational Linguistics. Meeting. vol. 2017. NIH Public Access; 2017. p. 299.

\bibitem{uzuner2010extracting}
Uzuner {\"O}, Solti I, Cadag E.
\newblock Extracting medication information from clinical text.
\newblock Journal of the American Medical Informatics Association. 2010;17(5):514-8.

\bibitem{henry20202018}
Henry S, Buchan K, Filannino M, Stubbs A, Uzuner O.
\newblock 2018 n2c2 shared task on adverse drug events and medication extraction in electronic health records.
\newblock Journal of the American Medical Informatics Association. 2020;27(1):3-12.

\bibitem{lybarger20232022}
Lybarger K, Yetisgen M, Uzuner {\"O}.
\newblock The 2022 n2c2/UW shared task on extracting social determinants of health.
\newblock Journal of the American Medical Informatics Association. 2023;30(8):1367-78.

\bibitem{uzuner2008identifying}
Uzuner {\"O}, Goldstein I, Luo Y, Kohane I.
\newblock Identifying patient smoking status from medical discharge records.
\newblock Journal of the American Medical Informatics Association. 2008;15(1):14-24.

\bibitem{uzuner2009recognizing}
Uzuner {\"O}.
\newblock Recognizing obesity and comorbidities in sparse data.
\newblock Journal of the American Medical Informatics Association. 2009;16(4):561-70.

\bibitem{stubbs2019cohort}
Stubbs A, Filannino M, Soysal E, Henry S, Uzuner {\"O}.
\newblock Cohort selection for clinical trials: n2c2 2018 shared task track 1.
\newblock Journal of the American Medical Informatics Association. 2019;26(11):1163-71.

\bibitem{jullien2024semeval}
Jullien M, Valentino M, Freitas A.
\newblock SemEval-2024 task 2: Safe biomedical natural language inference for clinical trials.
\newblock arXiv preprint arXiv:240404963. 2024.

\bibitem{razuvayevskaya2023clinical}
Razuvayevskaya O, Lopez I, Dunham I, Ochoa D.
\newblock Why clinical trials stop: the role of genetics.
\newblock medRxiv. 2023:2023-02.

\bibitem{MTSample}
Welcome to MTSamples; 2023.
\newblock Accessed: 2024-6-8.
\newblock \url{https://mtsamples.com/}.

\bibitem{VANMULLIGEN2012879}
{van Mulligen} EM, Fourrier-Reglat A, Gurwitz D, Molokhia M, Nieto A, Trifiro G, et~al.
\newblock The EU-ADR corpus: Annotated drugs, diseases, targets, and their relationships.
\newblock Journal of Biomedical Informatics. 2012;45(5):879-84.
\newblock Text Mining and Natural Language Processing in Pharmacogenomics.
\newblock Available from: \url{https://www.sciencedirect.com/science/article/pii/S1532046412000573}.

\bibitem{bastan2022bionli}
Bastan M, Surdeanu M, Balasubramanian N.
\newblock BioNLI: Generating a Biomedical NLI Dataset Using Lexico-semantic Constraints for Adversarial Examples.
\newblock In: Findings of the Association for Computational Linguistics: EMNLP 2022; 2022. p. 5093-104.

\bibitem{bowman2015large}
Bowman SR, Angeli G, Potts C, Manning CD.
\newblock A large annotated corpus for learning natural language inference.
\newblock arXiv preprint arXiv:150805326. 2015.

\bibitem{N18-1101}
Williams A, Nangia N, Bowman S.
\newblock A Broad-Coverage Challenge Corpus for Sentence Understanding through Inference.
\newblock In: Proceedings of the 2018 Conference of the North American Chapter of the Association for Computational Linguistics: Human Language Technologies, Volume 1 (Long Papers). Association for Computational Linguistics; 2018. p. 1112-22.
\newblock Available from: \url{http://aclweb.org/anthology/N18-1101}.

\bibitem{cohan-etal-2018-discourse}
Cohan A, Dernoncourt F, Kim DS, Bui T, Kim S, Chang W, et~al.
\newblock A Discourse-Aware Attention Model for Abstractive Summarization of Long Documents.
\newblock In: Proceedings of the 2018 Conference of the North {A}merican Chapter of the Association for Computational Linguistics: Human Language Technologies, Volume 2 (Short Papers). New Orleans, Louisiana: Association for Computational Linguistics; 2018. p. 615-21.
\newblock Available from: \url{https://aclanthology.org/N18-2097}.

\bibitem{guo2021automated}
Guo Y, Qiu W, Wang Y, Cohen T.
\newblock Automated lay language summarization of biomedical scientific reviews.
\newblock In: Proceedings of the AAAI Conference on Artificial Intelligence. vol.~35; 2021. p. 160-8.

\bibitem{yim2023aci}
Yim Ww, Fu Y, Ben~Abacha A, Snider N, Lin T, Yetisgen M.
\newblock Aci-bench: a novel ambient clinical intelligence dataset for benchmarking automatic visit note generation.
\newblock Scientific Data. 2023;10(1):586.

\bibitem{ainsworth2022git}
Ainsworth SK, Hayase J, Srinivasa S.
\newblock Git re-basin: Merging models modulo permutation symmetries.
\newblock arXiv preprint arXiv:220904836. 2022.

\bibitem{yu2023language}
Yu L, Yu B, Yu H, Huang F, Li Y.
\newblock Language models are super mario: Absorbing abilities from homologous models as a free lunch.
\newblock arXiv preprint arXiv:231103099. 2023.

\bibitem{suominen2013overview}
Suominen H, Salanter{\"a} S, Velupillai S, Chapman WW, Savova G, Elhadad N, et~al.
\newblock Overview of the ShARe/CLEF eHealth evaluation lab 2013.
\newblock In: Information Access Evaluation. Multilinguality, Multimodality, and Visualization: 4th International Conference of the CLEF Initiative, CLEF 2013, Valencia, Spain, September 23-26, 2013. Proceedings 4. Springer; 2013. p. 212-31.

\bibitem{wang2020medsts}
Wang Y, Afzal N, Fu S, Wang L, Shen F, Rastegar-Mojarad M, et~al.
\newblock MedSTS: a resource for clinical semantic textual similarity.
\newblock Language Resources and Evaluation. 2020;54:57-72.

\bibitem{chung2024scaling}
Chung HW, Hou L, Longpre S, Zoph B, Tay Y, Fedus W, et~al.
\newblock Scaling instruction-finetuned language models.
\newblock Journal of Machine Learning Research. 2024;25(70):1-53.

\bibitem{wu2024pmc}
Wu C, Lin W, Zhang X, Zhang Y, Xie W, Wang Y.
\newblock PMC-LLaMA: toward building open-source language models for medicine.
\newblock Journal of the American Medical Informatics Association. 2024:ocae045.

\bibitem{tunstall2023zephyr}
Tunstall L, Beeching E, Lambert N, Rajani N, Rasul K, Belkada Y, et~al.
\newblock Zephyr: Direct distillation of lm alignment.
\newblock arXiv preprint arXiv:231016944. 2023.

\bibitem{LLaMA3}
at~Meta A. Introducing Meta Llama 3: The most capable openly available LLM to date; 2024.
\newblock Accessed: 2024-04-18.
\newblock \url{https://ai.meta.com/blog/meta-llama-3/}.

\bibitem{kaplan2020scaling}
Kaplan J, McCandlish S, Henighan T, Brown TB, Chess B, Child R, et~al.
\newblock Scaling laws for neural language models.
\newblock arXiv preprint arXiv:200108361. 2020.

\bibitem{wu2011semantic}
Wu S, Liu H.
\newblock Semantic characteristics of NLP-extracted concepts in clinical notes vs. biomedical literature.
\newblock In: AMIA Annual Symposium Proceedings. vol. 2011. American Medical Informatics Association; 2011. p. 1550.

\bibitem{zhang2023promptner}
Zhang M, Yan H, Zhou Y, Qiu X.
\newblock Promptner: A prompting method for few-shot named entity recognition via k nearest neighbor search.
\newblock arXiv preprint arXiv:230512217. 2023.

\bibitem{gero2023self}
Gero Z, Singh C, Cheng H, Naumann T, Galley M, Gao J, et~al.
\newblock Self-verification improves few-shot clinical information extraction.
\newblock arXiv preprint arXiv:230600024. 2023.

\bibitem{lewis2020retrieval}
Lewis P, Perez E, Piktus A, Petroni F, Karpukhin V, Goyal N, et~al.
\newblock Retrieval-augmented generation for knowledge-intensive nlp tasks.
\newblock Advances in Neural Information Processing Systems. 2020;33:9459-74.

\end{thebibliography}

\end{document}